\documentclass[10pt,twocolumn,letterpaper]{article}

\usepackage[pagenumbers]{cvpr} % To force page numbers, e.g. for an arXiv version

\usepackage{multirow}
\usepackage{colortbl}

\definecolor{cvprblue}{rgb}{0.21,0.49,0.74}
\usepackage[pagebackref,breaklinks,colorlinks,allcolors=cvprblue]{hyperref}

\def\confName{CVPR}
\def\confYear{2027}

\title{EvoVAU: Tool Evolution for Video Anomaly Understanding}

\author{%
Benfeng Wang\textsuperscript{1}\quad Chao Huang\textsuperscript{1}\quad
Yudong Cheng\textsuperscript{2}\quad Chengliang Liu\textsuperscript{3}\quad
Wei Wang\textsuperscript{1}\quad Wenqi Ren\textsuperscript{1}\\
\textsuperscript{1}Sun Yat-sen University\qquad
\textsuperscript{2}Fudan University\qquad
\textsuperscript{3}University of Macau\\
{\small\ttfamily wangbf23@mail2.sysu.edu.cn\quad huangch253@mail.sysu.edu.cn\quad ydcheng25@m.fudan.edu.cn}\\
{\small\ttfamily liucl1996@163.com\quad wangwei29@mail.sysu.edu.cn\quad renwq3@mail.sysu.edu.cn}
}

\begin{document}
\maketitle
\begin{abstract}

Video anomaly understanding (VAU) requires reasoning about event progression, interactions, and the contextual conditions that distinguish abnormal behavior from common activity across diverse scenes. Although video agents enable multi-round analysis, the predefined analytical operations may not adequately satisfy these diverse reasoning demands. In this paper, we present EvoVAU, a framework that evolves a reusable tool library for a frozen video solver. Guided by task outcomes and execution trajectories, an external proposer identifies potential limitations in the library and refines it, retaining task experience as analytical operations that can be reused across questions. Specifically, Cross-Version Contrastive Feedback guides tool refinement by combining current errors with paired execution trajectories from historical corrections and regressions. Adoption-Aware Selection accepts a candidate library only when overall development-set accuracy improves and corrections outnumber regressions among questions on which the solver invokes new or modified tools. Using a small development set, EvoVAU alternates tool refinement with candidate evaluation and selection. At test time, the evolved library remains fixed, with no further involvement from the proposer. Experiments on several VAU benchmarks demonstrate strong performance and the flexibility of the evolved library across anomaly domains.

\end{abstract}    
\section{Introduction}
\label{sec:intro}

\begin{figure}[t]
    \centering
    \includegraphics[width=1\linewidth]{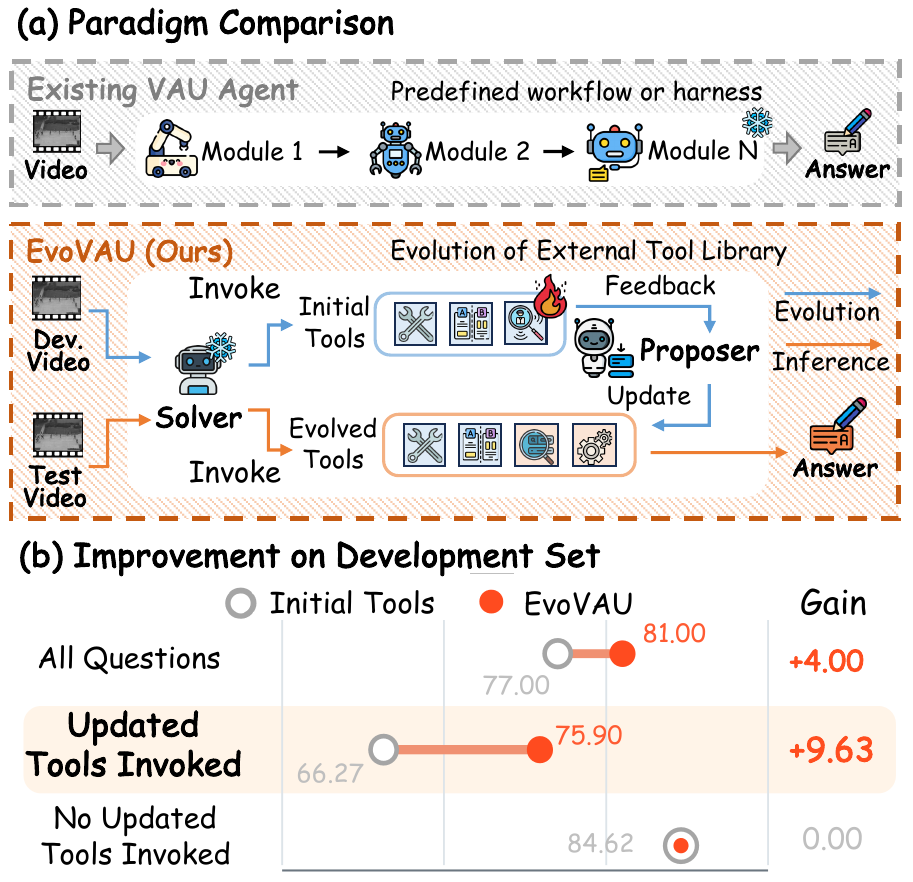}
    \caption{Motivation for EvoVAU.
    \textbf{(a) Paradigm comparison.} Existing VAU agents~\cite{panda-neurips2025,vto-acmmm2026,Anom-π-icml2026,tau-agent-arxiv2608} adapt analysis within predefined workflow or harness, whereas EvoVAU evolves tools for a frozen solver using task feedback and an external proposer.
    \textbf{(b) Improvement on the Development Set.} Initial tools and EvoVAU use the same frozen solver. Subsets are defined by whether the solver invokes updated tools.}
    \label{fig:motivation}
\end{figure}

Video anomaly understanding (VAU) aims to identify and interpret abnormal events in videos, supporting applications like safety monitoring and content moderation. In open-world settings, the diversity of scenes and events makes it difficult to recognize every form of abnormal behavior~\cite{hawk-neurips2025,openworld-vad-tmm2025}. VAU therefore requires reasoning about both how an event unfolds and why it is abnormal in its particular context. Multimodal Large Language Models (MLLMs) offer a promising foundation for VAU through their ability to describe events and reason about them in natural language~\cite{hawk-neurips2025,holmes-vau-cvpr2025,cuva-cvpr2024,vad-r1-neurips2025,target-vau-aaai2026}. Existing approaches often reduce video inputs through uniform frame sampling~\cite{hawk-neurips2025,vad-r1-neurips2025} or confidence-based frame selection within suspicious temporal regions~\cite{holmes-vau-cvpr2025,target-vau-aaai2026}. However, a limited set of frames may omit a brief action or the surrounding events needed to distinguish an anomaly from common activity. Effective VAU thus calls for analysis that adapts to the temporal and contextual demands of different abnormal events.

Recent agent-based approaches offer flexibility through task decomposition, collaboration among specialized agents, and iterative refinement of the analysis~\cite{panda-neurips2025,Anom-π-icml2026,vto-acmmm2026,tau-agent-arxiv2608,o-vad-eccv2026}. These methods allow a system to revisit relevant video intervals and adjust its reasoning as a task progresses. However, their analytical operations and coordination mechanisms are predefined in advance, as illustrated in Figure~\ref{fig:motivation}(a). Across domains, anomaly understanding may require reconstructing an incident's onset and consequences, examining interactions, or interpreting the contextual conditions that make an anomaly. A fixed set of capabilities may not support these demands equally well. For example, distinguishing a brief accident from routine motion may require examining the surrounding sequence, while judging whether an action violates a scene-specific restriction may require interpreting a visible warning. Such analytical needs can recur across otherwise different scenes and anomaly types. This motivates developing reusable operations from task experience, so that a video agent can expand its capabilities as well as adapt their invocation to different anomaly-related scenes.

In this paper, we propose EvoVAU, a tool evolution framework for VAU that couples a frozen video solver with an external proposer. Our core idea is to use feedback from task outcomes and execution trajectories on the development set to help the proposer identify potential shortcomings in the current tool library and refine it for VAU as shown in Figure~\ref{fig:motivation}(a). After tool refinement, candidate libraries are evaluated on development set and selectively retained, while the solver model remains frozen. Through tool refinement, EvoVAU enables the solver to select different analytical capabilities according to the specific scenes. The resulting capabilities are developed from limited task feedback and can be reused across scenes and anomaly types. After evolution, the solver uses the evolved tool library without further tool refinement or involvement from the proposer. As shown in Figure~\ref{fig:motivation}(b), this process improves overall development accuracy, with a pronounced gain on questions where EvoVAU invokes updated tools. In this run, that subset accounts for the entire development-set gain, while aggregate accuracy on questions without updated-tool invocation remains unchanged.

Achieving tool evolution requires feedback that guides tool refinement, together with a criterion for determining whether to accept each candidate library. To this end, we first introduce \textbf{Cross-Version Contrastive Feedback}, which combines current errors with paired execution trajectories from historical regressions and corrections. These experiences help the proposer infer potential shortcomings in the current tools and identify useful behaviors to preserve. Second, we introduce \textbf{Adoption-Aware Selection}, which accepts a candidate library only when overall development accuracy improves and corrections outnumber regressions on matched questions where the solver attempts to invoke refined tools. This criterion evaluates a candidate library through both overall task performance and the observed outcomes associated with the invocation of its refined tools. Our contributions are as follows:
\begin{itemize}
    \item We present EvoVAU, a tool evolution framework that couples a frozen video solver with an external proposer to refine tools for video anomaly understanding.
    \item We introduce Cross-Version Contrastive Feedback to guide an external proposer in refining tools for VAU from current errors and historical changes.
    \item We introduce Adoption-Aware Selection to control candidate acceptance through both overall task improvement and paired outcome improvements on questions where the solver attempts to invoke the candidate's refined tools.
    \item In-domain and cross-domain experiments demonstrate strong anomaly understanding performance and the flexibility of the evolved tool library across diverse tasks.
\end{itemize}

\section{Related Work}
\label{sec:related_work}

\subsection{Video Anomaly Detection and Understanding}
Traditional video anomaly detection (VAD) identifies and temporally localizes abnormal events through frame- or segment-level anomaly scores. Semi-supervised methods model normal patterns from normal training videos~\cite{self-spuervise-tcb2022,nwpu-tpami2024,prototype-ijcai2024}, while weakly supervised methods learn temporal anomaly scoring from video-level labels~\cite{vadclip-aaai2024,fed-wsvad-aaai2025,exvad-icml2025}. Multimodal large language models (MLLMs) extend anomaly detection toward event description, explanation, and question answering. HAWK~\cite{hawk-neurips2025} incorporates motion information to support open-world anomaly interpretation, while Holmes-VAU~\cite{holmes-vau-cvpr2025} combines hierarchical instruction data with anomaly-focused temporal sampling for understanding events at multiple temporal scales. Vad-R1~\cite{vad-r1-neurips2025} develops perception-to-cognition reasoning through structured chain-of-thought supervision and reinforcement learning. TargetVAU~\cite{target-vau-aaai2026} models spatiotemporal interactions among individuals to identify and explain anomalous target behaviors. Alongside these methods, CUVA~\cite{cuva-cvpr2024} broadens evaluation to anomaly descriptions, causes, and consequences. CueBench~\cite{cue-aaai2026} emphasizes the contextual conditions that distinguish abnormal events from normal ones. These studies establish increasingly detailed semantic objectives for VAU. However, under a limited frame budget, sampled observations may omit brief actions or the surrounding context needed to interpret them. Supporting analysis that adapts its temporal scope and contextual reasoning to different abnormal events therefore remains an important challenge.

\begin{figure*}[t]
    \centering
    \includegraphics[width=1\linewidth]{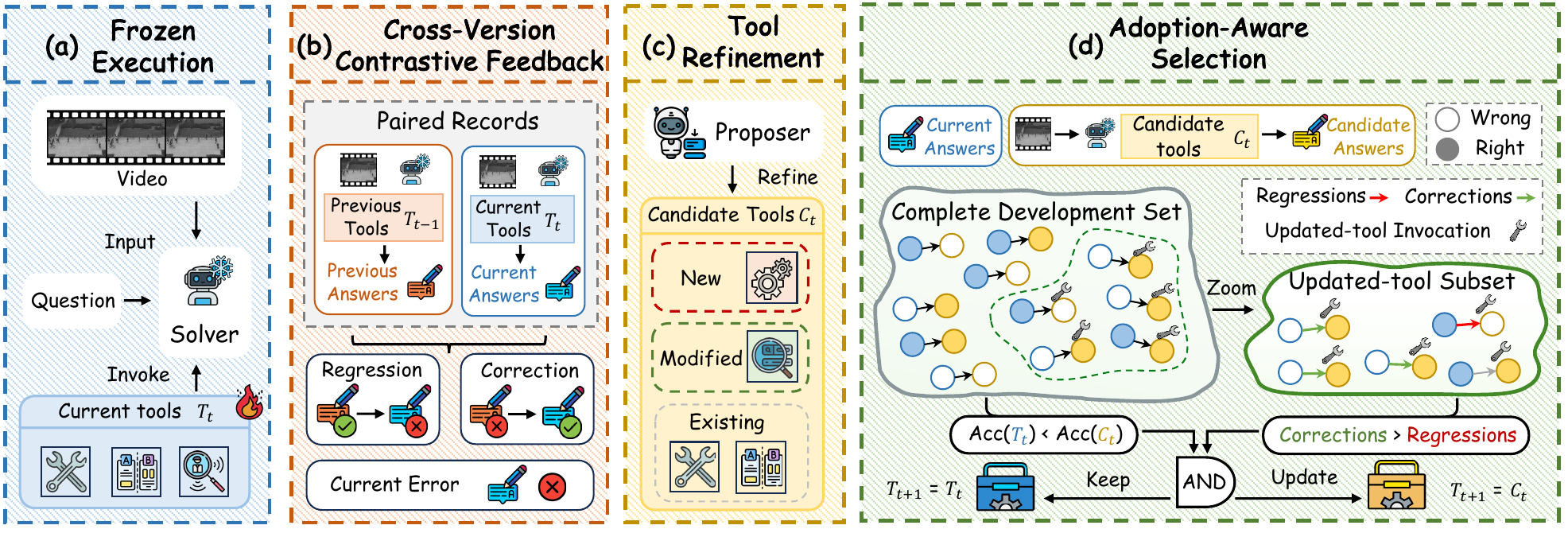}
    \caption{Overview of EvoVAU.
    \textbf{(a) Frozen Execution.} The video solver answers VAU questions using the current tool library $T_t$, with its model parameters frozen.
    \textbf{(b) Cross-Version Contrastive Feedback.} Paired execution records from the preceding round provide historical corrections and regressions, complemented by current errors.
    \textbf{(c) Tool Refinement.} The external proposer refines the tool library to form $C_t$, retaining unchanged tools.
    \textbf{(d) Adoption-Aware Selection.} The candidate is evaluated on the full development set. A candidate is accepted only if overall accuracy improves and corrections outnumber regressions on questions where the solver attempts to use a new or modified tool (wrench). Blue--yellow pairs compare current and candidate answers to the same question; filled and hollow nodes indicate correct and incorrect answers. Node counts are illustrative.}
    \label{fig:main}
\end{figure*}

\subsection{Agentic Video Understanding}

Agentic video understanding organizes information acquisition and reasoning into interactive procedures. VideoAgent~\cite{videoagent-eccv2024} iteratively gathers relevant information through visual description and retrieval tools. VideoMind~\cite{videomind-iclr2026} coordinates four subagents through role-specific LoRA adapters, while EVA~\cite{eva-arxiv2603} learns planning-before-perception behavior through supervised learning and reinforcement learning. LVAgent~\cite{lvagent-iccv2025} employs multiple MLLM agents through successive rounds of collaboration. 

For video anomaly analysis, PANDA~\cite{panda-neurips2025} combines scene-conditioned planning, tool-assisted reflection, and video-specific memory to refine detection decisions. Anom-$\pi$~\cite{Anom-π-icml2026} learns an interleaved reasoning and observation policy through interactive preference optimization. VTO~\cite{vto-acmmm2026} trains multi-step visual tool orchestration with process-level supervision. TAU-Agent~\cite{tau-agent-arxiv2608} retrieves captions and object trajectories to support a fine-tuned traffic-anomaly answering model, while O-VAD~\cite{o-vad-eccv2026} integrates object grounding, state tracking, and contextual reasoning with visual verification for industrial anomaly analysis. While these agents adapt inference within predefined workflows, the proposed EvoVAU uses task feedback to refine its tool library, enabling reuse of its tools for anomaly-related questions with different temporal and contextual demands.

\subsection{Self-Evolving Video Understanding Agents}

Self-evolving video agents improve their capabilities by iteratively learning from feedback. Model evolution~\cite{video-zero-arxiv2605,evo-ground-arxiv2605,evovid-arxiv2605} updates model parameters using supervision generated from unlabeled videos. These methods couple question generation with solver learning, emphasizing temporal grounding in the generated supervision or training objectives. Harness evolution~\cite{video-rsi-arxiv2609, videoharness-rsi-arxiv2608, vidharness-arxiv2609, videoevoharness-arxiv2610} optimizes the programs surrounding a frozen model, using feedback and candidate search to revise context construction and inference procedures, with validation based on development accuracy and, in some cases, computational cost. Tool evolution adapts reusable operations, such as macro-tools distilled from successful trajectories~\cite{meta-cvpr2026} and the joint evolution of policies and executable tools~\cite{coevowhen-arxiv2609}. The proposed EvoVAU focuses on VAU, using cross-version feedback to guide tool refinement and adoption-aware selection to assess candidate libraries through both overall performance and outcomes on questions where the solver attempts to use new or modified tools.

\section{Method}
\label{sec:method}

EvoVAU evolves the tool library for a frozen video solver through a loop of feedback construction, tool refinement, and candidate selection on a small development set, as shown in Figure~\ref{fig:main}.

\subsection{Task Formulation and Overview}
\label{sec:method_overview}

\paragraph{Task Formulation.}
We study VAU through multiple-choice questions about whether abnormal events occur, what they are, and how they unfold in context. Each sample $(V_i,q_i,y_i)$ consists of a video, a question with answer options, and the correct answer. Using a tool library $T$, the video solver $\mathcal S_\theta$ returns an answer and a trajectory:

\begin{equation}
    (\hat y_i(T),\tau_i(T))
    =\mathcal S_\theta(V_i,q_i;T).
    \label{eq:solver_execution}
\end{equation}

The trajectory $\tau_i(T)$ records intermediate interactions, including tool calls and returned observations. Each tool exposes a callable interface with a description and an input schema, backed by an executable implementation. Tools encapsulate video access, retrieval, or model-based analysis as reusable operations. The solver autonomously selects and invokes these operations as needed to examine event progression, interactions, and the context that distinguishes abnormal behavior from common activity. 

Different abnormal events can share an analytical demand, such as examining a brief transition or recovering the context of an interaction. A reusable tool preserves the procedure for addressing this demand and allows the same operation to support different event instances and domains. Tool evolution iteratively refines the library of reusable operations, while the frozen solver uses the evolved library to analyze abnormal events in different contexts. This separation supports flexible reuse of analytical capabilities across diverse anomaly understanding tasks.

\paragraph{Overview of Evolution.}

Let $\mathcal D=\{(V_i,q_i,y_i)\}_{i=1}^{N}$ be the development set, with question indices $i\in\{1,\ldots,N\}$. For any tool library $T$, we define the correctness indicator as $c_i(T)=\mathbf{1}[\hat y_i(T)=y_i]$, where $\mathbf{1}[\cdot]$ equals one when its condition holds and zero otherwise. Let $\mathbb T$ denote the space of executable tool libraries. EvoVAU seeks to maximize development accuracy by optimizing $T$ while keeping the solver parameters $\theta$ frozen:
\begin{equation}
    \begin{aligned}
        \operatorname{Acc}(T)&=\frac{1}{N}\sum_{i=1}^{N}c_i(T),\\
        T^{\star}&\in\arg\max_{T\in\mathbb T}\operatorname{Acc}(T).
    \end{aligned}
    \label{eq:tool_objective}
\end{equation}

Starting from an initial library $T_0$, EvoVAU iterates through the four stages in Figure~\ref{fig:main}. At iteration $t$, the frozen solver answers development questions using $T_t$, yielding predictions and trajectories, as shown in Figure~\ref{fig:main}(a). Comparing its answers with those obtained using the previous library $T_{t-1}$ provides historical corrections and regressions. Together with current errors, they form the Cross-Version Contrastive Feedback detailed in Section~\ref{sec:contrastive_feedback}. Using this feedback, the external proposer $\mathcal P$ refines $T_t$ into a candidate library $C_t$ by creating new tools or modifying existing ones, as illustrated in Figure~\ref{fig:main}(c). Following the Adoption-Aware Selection process described in Section~\ref{sec:adoption_selection}, we evaluate $C_t$ on the full development set. If overall accuracy improves and corrections outnumber regressions on questions where its refined tools are invoked, $C_t$ is accepted as $T_{t+1}$. Otherwise, $T_t$ is retained. The final library is fixed for test-time inference, without participation from the proposer.

\subsection{Cross-Version Contrastive Feedback}
\label{sec:contrastive_feedback}

The frozen solver answers VAU questions using the current tool library. The resulting trajectories record its tool calls and observations. An incorrect answer may reflect a failure to recover an event's development or the contextual conditions that distinguish abnormal behavior from an apparently similar common action. The trajectory leading to an incorrect answer provides limited guidance on which alternative analysis might help. We therefore compare executions of the same question across tool versions and combine historical corrections and regressions with current errors, as illustrated in Figure~\ref{fig:main}(b). This feedback provides examples of useful analytical behavior, previously successful behavior to preserve, and unresolved demands for tool refinement.

\paragraph{Historical Corrections and Regressions.}
Using paired records from the previous and current tool libraries, we define the correction set $\mathcal H_t^{+}$ and regression one $\mathcal H_t^{-}$ as

\begin{equation}
    \begin{aligned}
        \mathcal H_t^{+}&=\{i\in\{1,\ldots,N\}:c_i(T_{t-1})=0 \land c_i(T_t)=1\},\\
        \mathcal H_t^{-}&=\{i\in\{1,\ldots,N\}:c_i(T_{t-1})=1 \land c_i(T_t)=0\}.
    \end{aligned}
    \label{eq:historical_changes}
\end{equation}

Here, $\mathcal H_t^{+}$ contains questions whose answers change from incorrect with $T_{t-1}$ to correct with $T_t$, while $\mathcal H_t^{-}$ captures the reverse transition. For each historical case, the feedback pairs the two predicted answers and trajectories for the same question with its reference answer. The proposer can examine how the analytical operations and their outputs changed alongside correctness. Corrections suggest useful ways to analyze abnormal events, while regressions highlight previously successful behavior to preserve.

\paragraph{Current Errors.}
Questions with $c_i(T_t)=0$ expose demands that the current library has yet to address. Their answers and trajectories, together with the questions and references, help the proposer inspect how the solver approached the unresolved anomaly interpretation. These cases complement historical changes by highlighting remaining difficulties, even when no change in correctness has occurred.

Overall, the value of a historical pair lies in comparing different analyses of the same anomaly-related question. With the video, question, and reference answer held fixed, the proposer can inspect changes in the temporal context examined, interactions considered, and the sequence of analytical operations. Corrections and regressions provide complementary examples of behavior to investigate, while current errors keep tool refinement focused on unresolved tasks. This combination helps the proposer consider both remaining analytical needs and previously useful behavior. 

Using this feedback, the external proposer identifies analytical demands that emerge from the solver's task experience and are not adequately supported by the current tool library. As shown in Figure~\ref{fig:main}(c), it refines the tool library in response to these findings, forming the candidate library $C_t$ while retaining unchanged tools.

\subsection{Adoption-Aware Selection}
\label{sec:adoption_selection}

The candidate library $C_t$ is then evaluated on the full development set with the same frozen solver. The solver may invoke the candidate's refined tools for only a subset of questions. Higher overall accuracy alone does not establish whether these questions benefit from tool refinement. As illustrated in Figure~\ref{fig:main}(d), we therefore compare the current library $T_t$ and candidate one $C_t$ both on the full development set and on the questions where the solver uses the candidate's refined tools.

\paragraph{Updated-Tool Invocation.}

Let $\Delta T_t$ denote the set of refined tools, i.e., tools newly created or modified in $C_t$ relative to $T_t$. For an execution trajectory $\tau$, $\operatorname{Calls}(\tau)$ denotes the set of tools that the solver attempts to invoke in trajectory $\tau$. We define the corresponding subset of questions as

\begin{equation}
    U_t=\{i\in\{1,\ldots,N\}:\operatorname{Calls}(\tau_i(C_t))
    \cap\Delta T_t\neq\varnothing\}.
    \label{eq:adopted_subset}
\end{equation}

Thus, $i\in U_t$ means that the solver attempts to invoke at least one refined tool from $\Delta T_t$ when answering question $i$ with the candidate library $C_t$. 

\paragraph{Paired Outcome Comparison.}

Following the same definition of outcome transitions as in Section~\ref{sec:contrastive_feedback}, we define corrections and regressions within $U_t$ as

\begin{equation}
    \begin{aligned}
        U_t^{+}&=\{i\in U_t:c_i(T_t)=0 \land c_i(C_t)=1\},\\
        U_t^{-}&=\{i\in U_t:c_i(T_t)=1 \land c_i(C_t)=0\}.
    \end{aligned}
    \label{eq:adopted_outcomes}
\end{equation}

A candidate is accepted only when it improves overall accuracy and yields more corrections than regressions within the subset $U_t$

\begin{equation}
    \operatorname{Acc}(C_t)>\operatorname{Acc}(T_t)
    \quad\land\quad |U_t^{+}|>|U_t^{-}|.
    \label{eq:acceptance}
\end{equation}

Overall accuracy measures the performance of the complete system, while accuracy on $U_t$ provides evidence of the effectiveness of refined tools when invoked. Tool refinement may also alter the solver's use of unchanged tools, affecting outcomes on questions outside $U_t$. Consequently, an overall accuracy gain may come from these other questions even when accuracy within $U_t$ does not improve. Conversely, gains within $U_t$ may be outweighed by losses elsewhere, reducing overall accuracy. We therefore require both improved overall accuracy and a positive net gain within $U_t$ before accepting a candidate library.

\section{Experiments}
\label{sec:experiments}

% We evaluate EvoVAU on in-domain and out-of-domain VAU benchmarks, followed by ablations of its feedback and selection mechanisms and comparisons before and after tool evolution.

% Source: https://my.feishu.cn/wiki/XZPqwS1USihq61kEPr1ckRdInKg
% Sheet: 主要结果1 (ziBwYE), revision 286, retrieved 2026-11-08.
% Qwen and Ovis rows are excluded as requested. The helper max row is not a method.
% All overall scores are preserved from the source; best values are computed over included methods.
% Requires booktabs, already loaded by cvpr.sty.
\begin{table}[t]
    \caption{In-domain results on VAU-Bench~\cite{vau-r1-arxiv2505}. * The results of VAU-R1 are from the original paper~\cite{vau-r1-arxiv2505}.}
    \label{tab:result_in_domain}
    \centering
    \small
    \setlength{\tabcolsep}{8pt}
    \begin{tabular}{l c c c}
        \toprule
        Method & QA & CLS & Binary \\
        \midrule
        \rowcolor{gray!20}
        \multicolumn{4}{l}{\textit{General models}} \\
        \midrule
        % Source row 4.
        InternVL3.5-8B~\cite{internvl35-arxiv2508} & 92.36 & 52.74 & 77.50 \\
        % Source row 5.
        MiMo-VL-7B~\cite{mimo-arxiv2506} & 91.50 & 56.08 & \textbf{82.99} \\
        % Source row 6.
        Keye-VL-1.5-8B~\cite{keye15-arxiv2509} & 87.19 & 49.41 & 71.26 \\
        % Source row 7.
        GLM-4.1V-9B~\cite{glm-arxiv2507} & 87.41 & 46.72 & 79.22 \\
        % Source row 9.
        LLaVA-OneVision-7B~\cite{llava-ov-arvix2408} & 86.01 & 46.50 & 64.69 \\
        \midrule
        \rowcolor{gray!20}
        \multicolumn{4}{l}{\textit{Video agents}} \\
        \midrule
        % Source row 13.
        VideoAgent~\cite{videoagent-eccv2024} & 87.51 & 45.86 & 65.55 \\
        % Source row 14.
        VideoMind~\cite{videomind-iclr2026} & 89.67 & 48.76 & 78.36 \\
        % Source row 15.
        EVA~\cite{eva-arxiv2603} & 85.76 & 47.26 & 77.72 \\
        \midrule
        \rowcolor{gray!20}
        \multicolumn{4}{l}{\textit{Specialized models}} \\
        \midrule
        % Source row 18.
        HAWK~\cite{hawk-neurips2025} & 58.45 & 22.28 & 61.03 \\
        % Source row 19.
        HolmesVAU~\cite{holmes-vau-cvpr2025} & 77.72 & 33.69 & 62.00 \\
        % Source row 20.
        HolmesVAD~\cite{holmes-vad-arxiv2406} & 73.52 & 26.37 & 30.79 \\
        % Source row 21.
        Vad-R1~\cite{vad-r1-neurips2025} & 90.10 & 54.68 & 81.70 \\
        % Source row 22.
        Cue-R1~\cite{cue-aaai2026} & 86.33 & 39.94 & 73.20 \\
        % Source row 23.
        VAU-R1*~\cite{vau-r1-arxiv2505} & 88.04 & 45.76 & 48.45 \\
        \midrule
        % Source row 25.
        \textbf{EvoVAU} & \textbf{93.33} & \textbf{56.30} & 79.87 \\
        \bottomrule
    \end{tabular}
\end{table}

\subsection{Experimental Setup}
\label{sec:experimental_setup}

\paragraph{Datasets and Benchmarks.}

We sample 200 question--answer pairs from the VAU-Bench~\cite{vau-r1-arxiv2505} training set to construct the development set for tool evolution. We evaluate in-domain performance on the complete VAU-Bench test set, which covers multiple-choice question answering (QA), anomaly classification (CLS), and binary classification. We also evaluate cross-domain performance on VANE-Bench~\cite{vane-naacl2025}, VRU-Accident~\cite{vru-iccvw2025}, and VALU~\cite{valu-acl2026} with accuracy. All questions use a multiple-choice format. The final tool library is fixed during evaluation, with no further tool refinement on the evaluation benchmarks. Please refer to Supplementary Materials for more information about datasets and benchmarks.

\begin{figure}[t]
    \centering
    \includegraphics[width=1\linewidth]{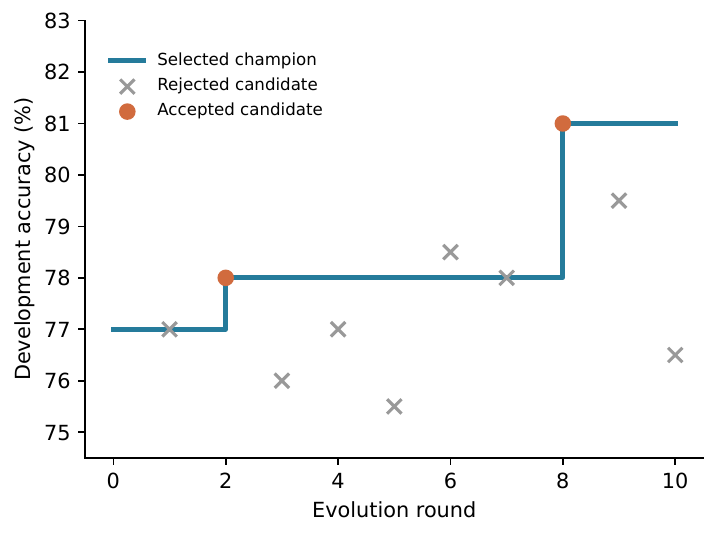}
    \caption{Development-set accuracy during tool evolution. The step curve tracks the retained library, while the circles and crosses denote accepted and rejected candidates.}
    \label{fig:evolution}
\end{figure}

\begin{table*}[t]
    \caption{Out-of-domain results on VANE-Bench~\cite{vane-naacl2025}, VRU-Accident~\cite{vru-iccvw2025}, and VALU~\cite{valu-acl2026}. The values are accuracy. Real and AI denote real and AI-generated videos, respectively.}
    \label{tab:result_out_domian}
    \centering
    \small
    \setlength{\tabcolsep}{5pt}
    \begin{tabular}{l c c c c c c c c c c c c}
        \toprule
        % Center the first heading across the two header rows.
        \multirow{2}{*}{Method} & \multicolumn{3}{c}{VANE-Bench} & \multicolumn{5}{c}{VRU-Accident} & \multicolumn{4}{c}{VALU} \\
        \cmidrule(lr){2-4}\cmidrule(lr){5-9}\cmidrule(l){10-13}
        & Real & AI & Avg. & CAP & DADA & DoTA & Manual & Avg. & UCF & MSAD & ECVA & Avg. \\
        \midrule
        \rowcolor{gray!20}
        \multicolumn{13}{l}{\textit{General models}} \\
        \midrule
        % Source row 4.
        InternVL3.5-8B~\cite{internvl35-arxiv2508} & 78.35 & 85.48 & 83.01 & 51.63 & 50.37 & 49.00 & 47.31 & 49.40 & 45.42 & 45.61 & 43.25 & 44.62 \\
        % Source row 5.
        MiMo-VL-7B~\cite{mimo-arxiv2506} & 79.90 & 72.88 & 75.31 & 52.21 & 48.95 & 46.00 & 48.29 & 49.33 & 47.12 & 54.01 & 48.77 & 50.40 \\
        % Source row 6.
        Keye-VL-1.5-8B~\cite{keye15-arxiv2509} & 79.38 & 77.53 & 78.18 & 60.16 & 58.97 & 56.17 & 59.36 & 59.18 & 23.98 & 35.27 & 34.15 & 32.36 \\
        % Source row 7.
        GLM-4.1V-9B~\cite{glm-arxiv2507} & 73.20 & 68.77 & 70.30 & 51.80 & 51.49 & 47.17 & 46.92 & 49.37 & 56.28 & 57.64 & 61.45 & 58.88 \\
        % Source row 9.
        LLaVA-OneVision-7B~\cite{llava-ov-arvix2408} & 68.04 & 47.12 & 54.38 & 66.03 & 62.56 & 58.83 & 62.52 & 63.17 & 48.25 & 49.67 & 53.17 & 50.77 \\
        \midrule
        \rowcolor{gray!20}
        \multicolumn{13}{l}{\textit{Video agents}} \\
        \midrule

        % Source row 13.
        VideoAgent~\cite{videoagent-eccv2024} & 68.04 & 68.49 & 68.34 & 67.19 & 66.29 & 66.50 & 66.62 & 66.70 & 41.74 & 41.12 & 40.59 & 41.04 \\
        % Source row 14.
        VideoMind~\cite{videomind-iclr2026} & 78.35 & 70.96 & 73.52 & 63.70 & 63.15 & 61.50 & 63.42 & 63.25 & 48.35 & 52.38 & 53.37 & 51.90 \\
        % Source row 15.
        EVA~\cite{eva-arxiv2603} & 69.59 & 63.01 & 65.30 & 68.29 & 60.31 & 65.67 & 64.40 & 64.73 & 18.60 & 23.40 & 23.42 & 22.36 \\
        % Source row 16.
        VideoHarness-RSI~\cite{videoharness-rsi-arxiv2608} & \textbf{83.51} & 73.70 & 77.10 & 69.57 & 66.07 & 65.50 & 66.58 & 67.22 & 52.79 & 52.82 & 55.47 & 53.88 \\
        \midrule
        \rowcolor{gray!20}
        \multicolumn{13}{l}{\textit{Specialized models}} \\
        \midrule

        % Source row 18.
        HAWK~\cite{hawk-neurips2025} & 5.67 & 17.26 & 13.24 & 37.05 & 34.60 & 35.67 & 34.53 & 35.38 & 14.64 & 18.74 & 18.40 & 17.71 \\
        % Source row 19.
        HolmesVAU~\cite{holmes-vau-cvpr2025} & 53.09 & 38.08 & 43.29 & 54.94 & 58.00 & 50.83 & 53.03 & 54.47 & 21.44 & 24.59 & 28.94 & 25.65 \\
        % Source row 20.
        HolmesVAD~\cite{holmes-vad-arxiv2406} & 15.98 & 24.93 & 21.82 & 41.17 & 42.97 & 38.00 & 38.85 & 40.35 & 0.57 & 0.65 & 1.48 & 0.97 \\
        % Source row 21.
        Vad-R1~\cite{vad-r1-neurips2025} & 76.29 & 72.05 & 73.52 & 59.93 & 55.23 & 52.17 & 55.21 & 56.27 & 57.32 & 59.21 & \textbf{61.55} & 59.74 \\
        % Source row 22.
        Cue-R1~\cite{cue-aaai2026} & 70.62 & 76.71 & 74.60 & 58.13 & 54.86 & 54.17 & 55.26 & 55.88 & 36.64 & 39.38 & 43.40 & 40.40 \\
        \midrule
        % Source row 25.
        \textbf{EvoVAU} & 76.29 & \textbf{87.40} & \textbf{83.54} & \textbf{74.10} & \textbf{69.28} & \textbf{69.50} & \textbf{70.51} & \textbf{71.17} & \textbf{60.81} & \textbf{65.28} & 61.09 & \textbf{62.62} \\
        \bottomrule
    \end{tabular}
\end{table*}

\paragraph{Implementation Details.}

Our main experiments use Qwen3.8-27B~\cite{qwen38} as the solver model if not specified. And we additionally use models with different architectures and sizes, including Qwen3.5-9B~\cite{qwen35}, ERNIE-4.5-VL-28B-A3B~\cite{ernie}, Gemma-4-31B-it, Gemma-4-26B-A4B~\cite{gemma4}, and InternVL3.5-30B-A3B~\cite{internvl35-arxiv2508}. We disable the solver's thinking mode and set its decoding temperature to 0. The external proposer is GPT-5.6-Sol~\cite{gpt56} through Codex~\cite{codex}, with reasoning effort set to high. We also provide access to a frozen EmbeddingGemma~2~\cite{embeddinggemma-arxiv2610} model for multimodal representation and similarity-based matching. During evolution, only the executable tool library is refined, while the solver and embedding model remain frozen. The proposer participates only in offline evolution. Further implementation details are provided in the Supplementary Materials.

\paragraph{Baselines.}

% different size?
We compare EvoVAU with three categories of models, including general MLLMs~\cite{internvl35-arxiv2508,mimo-arxiv2506,keye15-arxiv2509,glm-arxiv2507,llava-ov-arvix2408}, anomaly-specialized MLLMs~\cite{hawk-neurips2025,holmes-vad-arxiv2406,holmes-vau-cvpr2025,vad-r1-neurips2025,cue-aaai2026,vau-r1-arxiv2505}, and video agents~\cite{videoagent-eccv2024,videomind-iclr2026,eva-arxiv2603}. The general and anomaly-specialized MLLMs receive 32 frames per video. We also evaluate the concurrent VideoHarness-RSI~\cite{videoharness-rsi-arxiv2608}, which evolves harness code for video understanding.

\subsection{Main Results}
\label{sec:main_results}

\paragraph{In-Domain Evaluation.}
As shown in Table~\ref{tab:result_in_domain}, EvoVAU achieves the highest QA and CLS accuracy among the compared methods on VAU-Bench~\cite{vau-r1-arxiv2505}, while remaining competitive performance on binary classification. Compared with VAU-R1~\cite{vau-r1-arxiv2505}, which performs reinforcement learning training, EvoVAU exceeds the reported VAU-R1 results across all three tasks while keeping the model frozen with a small scale development set. The advantage extends to other anomaly-specialized models.  EvoVAU substantially outperforms HAWK~\cite{hawk-neurips2025}, HolmesVAD~\cite{holmes-vad-arxiv2406}, HolmesVAU~\cite{holmes-vau-cvpr2025} and Cue-R1~\cite{cue-aaai2026} across all three tasks. It also surpasses Vad-R1~\cite{vad-r1-neurips2025}, the strongest specialized baseline on both QA and CLS. This performance highlights EvoVAU's strength in answering anomaly-related questions and distinguishing anomaly categories. Together, these results demonstrate that a frozen solver equipped with evolved tools can achieve strong anomaly understanding using limited task-specific development data.

% Source: https://my.feishu.cn/wiki/XZPqwS1USihq61kEPr1ckRdInKg
% Sheet: 消融实验 (3yVaTr), revision 225, retrieved 2026-11-08.
% Definitions and Qwen3.5-9B agent confirmed by the author on 2026-11-08.
% Source mapping: L -> base; L+U -> +Adoption-Aware Selection;
% C -> +Cross-Version Contrastive Feedback; C+U -> EvoVAU.
% Source overall values are preserved, not recomputed as unweighted means.
% Requires booktabs (cvpr.sty) and multirow (preamble.tex).
\begin{table*}[t]
    \caption{Ablation of Cross-Version Contrastive Feedback and Adoption-Aware Selection with Qwen3.5-9B~\cite{qwen35}. The base evolves tools using current-round feedback and accepts candidates only when overall development accuracy improves. The two variants independently add one mechanism. EvoVAU combines both strategies.}
    \label{tab:ablation}
    \centering
    \small
    \setlength{\tabcolsep}{9pt}
    \begin{tabular}{l c c c c c c c c}
        \toprule
        % Center the first heading across the two header rows.
        \multirow{2}{*}{Strategy} & \multicolumn{5}{c}{VRU-Accident~\cite{vru-iccvw2025}} & \multicolumn{3}{c}{VANE-Bench~\cite{vane-naacl2025}} \\
        \cmidrule(lr){2-6}\cmidrule(l){7-9}
        & CAP & DADA & DoTA & Manual & Avg. & Real & AI & Avg. \\
        \midrule
        % Source row 4: L.
        base & 67.31 & 60.46 & 63.83 & 64.15 & 64.20 & 59.79 & \textbf{68.49} & 65.47 \\
        % Source row 5: L+U.
        + Adoption-Aware Selection & 67.54 & 63.45 & 66.67 & 63.80 & 65.08 & 57.73 & 65.75 & 62.97 \\
        % Source row 6: C.
        + Cross-Version Contrastive Feedback & 65.51 & 59.12 & 62.67 & 59.83 & 61.58 & 61.86 & 67.12 & 65.30 \\
        \midrule
        % Source row 7: C+U.
        \textbf{EvoVAU} & \textbf{68.00} & \textbf{64.28} & \textbf{67.00} & \textbf{64.15} & \textbf{65.57} & \textbf{64.43} & 67.95 & \textbf{66.73} \\
        \bottomrule
    \end{tabular}
\end{table*}

\paragraph{Cross-Domain Evaluation.}

Table~\ref{tab:result_out_domian} shows the cross-domain results of EvoVAU which is developed using VAU-Bench~\cite{vau-r1-arxiv2505} without further tool refinement on the other target benchmarks. EvoVAU achieves the strongest aggregate performance on VANE-Bench~\cite{vane-naacl2025}, VRU-Accident~\cite{vru-iccvw2025}, and VALU~\cite{valu-acl2026}. On VRU-Accident, its advantage extends across all component subsets, showing that the aggregate result is supported by consistent performance across the evaluated data sources. On VANE-Bench and VALU, EvoVAU achieves the strongest overall performance, demonstrating the cross-domain utility of the evolved library. These results demonstrate the reuse of operations developed on VAU-Bench~\cite{vau-r1-arxiv2505} across anomaly domains. The tool library therefore supports anomaly analysis beyond its development setting without requiring further tool refinement for each evaluation domain.

\subsection{Ablation Studies}
\label{sec:ablations}

\paragraph{Feedback and Selection Strategies.}

We evaluate the effectiveness of the proposed feedback and selection strategies in Table~\ref{tab:ablation}. The evolution baseline uses current-round feedback without historical comparisons and selects candidates only by overall development accuracy. The two variants independently add Adoption-Aware Selection or Cross-Version Contrastive Feedback to this base, while EvoVAU combines both mechanisms. EvoVAU achieves the highest aggregate accuracy, reaching 65.57\% and 66.73\%. Compared with cross-version contrastive feedback alone, combining both mechanisms improves these scores by 3.99\% and 1.43\%. Compared with adoption-aware selection alone, the gains are 0.49\% and 3.76\%. The two mechanisms control different stages. The feedback guides tool updating, while selection determines whether the candidate will be kept. Combining the two mechanisms supports coordinating tool refinement and candidate acceptance.

\begin{table}[t]
    \caption{Comparison of inference and adaptation strategies on VRU-Accident~\cite{vru-iccvw2025} with same development set.}
    \label{tab:optimization_strategies}
    \centering
    \small
    \setlength{\tabcolsep}{6pt}
    \begin{tabular}{l c c c c c}
        \toprule
        Strategy & CAP & DADA & DoTA & Manual & Avg. \\
        \midrule
        \rowcolor{gray!20}
        \multicolumn{6}{l}{\textit{Qwen3.5-9B~\cite{qwen35}}} \\
        \midrule
        General & 65.56 & 63.30 & 59.67 & 61.20 & 62.77 \\
        SFT & 48.84 & 47.01 & 51.33 & 50.47 & 49.32 \\
        Initial tools & 65.04 & 60.24 & 60.33 & 61.24 & 62.02 \\
        \textbf{EvoVAU} & \textbf{68.00} & \textbf{64.28} & \textbf{67.00} & \textbf{64.15} & \textbf{65.57} \\
        \midrule
        \rowcolor{gray!20}
        \multicolumn{6}{l}{\textit{Qwen3.8-27B~\cite{qwen38}}} \\
        \midrule
        General & 72.24 & 67.04 & 65.67 & 69.87 & 69.50 \\
        Initial tools & 73.17 & 66.29 & 65.33 & 67.52 & 68.65 \\
        \textbf{EvoVAU} & \textbf{74.10} & \textbf{69.28} & \textbf{69.50} & \textbf{70.51} & \textbf{71.17} \\
                \midrule
        \rowcolor{gray!20}
        \multicolumn{6}{l}{\textit{VideoHarness-RSI~\cite{videoharness-rsi-arxiv2608}}} \\
        \midrule
        Initial & \textbf{70.56} & \textbf{66.97} & 67.00 & \textbf{67.22} & \textbf{68.10} \\
        Evolved & 69.74 & 66.52 & \textbf{68.50} & 66.97 & 67.82 \\
        \bottomrule
    \end{tabular}
\end{table}

\paragraph{Inference and Adaptation Strategies.}

Table~\ref{tab:optimization_strategies} compares inference and adaptation strategies on VRU-Accident~\cite{vru-iccvw2025}. The General means performing anomaly understanding with uniform sampling, while SFT fine-tunes Qwen3.5-9B on the development set. With both solvers, EvoVAU improves over Initial tools and outperforms General, demonstrating the benefit of evolved agentic inference over direct understanding of uniformly sampled frames. For Qwen3.5-9B, SFT substantially reduces accuracy across all subsets relative to General, whereas EvoVAU uses the same limited data to improve performance without updating model parameters. We additionally evolve VideoHarness-RSI~\cite{videoharness-rsi-arxiv2608} in same settings, which shows overall performance degradation. EvoVAU refines the tool library to support flexible task-specific analysis, with the solver selecting tools as needed for each question. With Qwen3.8-27B, EvoVAU also exceeds the evolved harness across all four subsets. Together, the comparisons with SFT and harness evolution support tool-library evolution as an effective adaptation strategy for VAU under limited development data.

\paragraph{Evolution across Solver Models.}

We further apply the same tool-evolution procedure to Gemma~\cite{gemma4}, ERNIE~\cite{ernie}, and InternVL~\cite{internvl35-arxiv2508} families. As shown in Table~\ref{tab:solver_models}, EvoVAU improves over the initial tool library with all four additional solvers. These consistent gains demonstrate the effectiveness of our tool-evolution strategy across multiple base models from different families.

\subsection{Discussion}
\label{sec:discussion}

% We analyze how the tool library evolves with the frozen Qwen3.8-27B solver, how the resulting operations are reused, and how accuracy changes across question groups.

\begin{figure}[t]
    \centering
    \includegraphics[width=1\linewidth]{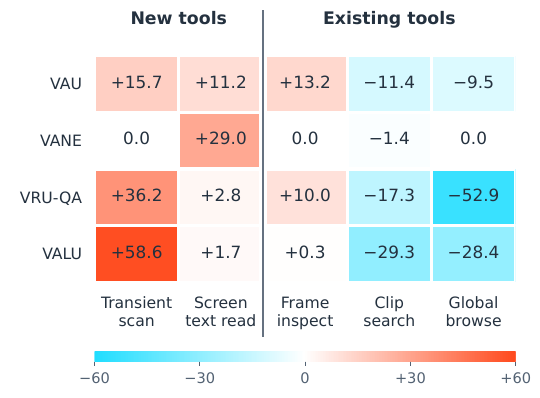}
    \caption{Changes in tool invocation coverage between EvoVAU and initial tools. Orange-red and cyan indicate increases and decreases, respectively.}
    \label{fig:usage}
\end{figure}

\begin{figure}[t]
    \centering
    \includegraphics[width=1\linewidth]{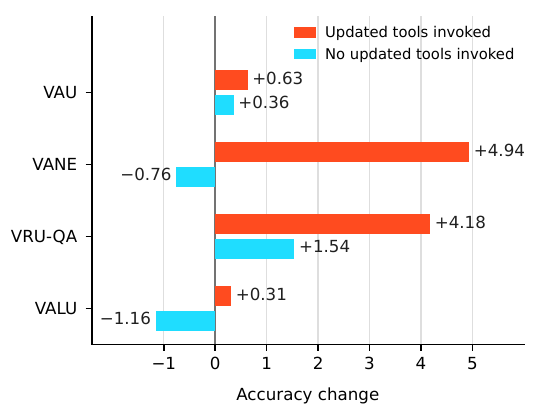}
    \caption{Accuracy changes from initial tools to EvoVAU within groups defined by updated-tool invocation.}
    \label{fig:change}
\end{figure}

\paragraph{Tool Evolution Process.}

As shown in Figure~\ref{fig:evolution}, candidate accuracy fluctuates across evolution rounds, with only two candidates accepted. Selection does not simply retain every candidate with higher overall accuracy. At round~6, the candidate improves overall accuracy but fails adoption-aware selection because corrections do not outnumber regressions within the subset $U_t$. The two accepted candidates introduce two new tools and raise development accuracy from 77\% to 81\%. Detailed analysis of the evolved tools is provided in the supplementary material.

\paragraph{Usage of Evolved Tools.}

We further analyze changes in tool usage before and after evolution, as shown in Figure~\ref{fig:usage}. Tool invocation coverage is defined as the percentage of questions with at least one attempted call to a tool. New tools have zero coverage under initial tools. On the one hand, the two new tools exhibit distinct coverage patterns across benchmarks: transient-scan covers a larger proportion of questions on VALU and VRU-QA, whereas screen-text-read has higher coverage on VANE and VAU. With the library fixed during evaluation, these different usage patterns show that a single evolved library can be reused across anomaly domains while allowing tool selection to vary with the task. The original tools also remain in use. On the other hand, the clip-search and global-browse coverage decline, while frame-inspection coverage increases a lot on VAU and VRU-QA. These trends suggest that new tools complement existing capabilities: specialized operations can address particular analytical needs, while the original tools continue to support observation and inspection of abnormal events. The expanded library gives the solver more ways to examine an event and its context.

\begin{table}[t]
    \caption{Tool evolution across additional solvers on VRU-Accident. $\Delta$ denotes the gain after evolution.}
    \label{tab:solver_models}
    \centering
    \small
    \setlength{\tabcolsep}{4pt}
    \begin{tabular}{l c c c}
        \toprule
        Solver & Initial tools & EvoVAU & $\Delta$ \\
        \midrule
        Gemma-4-31B~\cite{gemma4} & 66.17 & \textbf{66.70} & $+0.53$ \\
        Gemma-4-26B-A4B~\cite{gemma4} & 62.68 & \textbf{65.25} & $+2.57$ \\
        ERNIE-4.5-VL-28B-A3B~\cite{ernie} & 51.18 & \textbf{56.90} & $+5.72$ \\
        InternVL3.5-30B-A3B~\cite{internvl35-arxiv2508} & 56.78 & \textbf{59.45} & $+2.67$ \\
        \bottomrule
    \end{tabular}
\end{table}

\paragraph{Accuracy Changes by Tool Invocation.}

We extend the analysis of Figure~\ref{fig:motivation}(b) to evaluation benchmarks. As shown in Figure~\ref{fig:change}, we compare initial tools and EvoVAU within groups defined by updated-tool invocation. On the one hand, the accuracy of the updated-tool group improves on all four benchmarks, with the largest gain of 4.94\% on VANE-Bench. These gains suggest that the updated tools provide additional analytical capabilities that help address limitations of the initial tool library. On the other hand, the group without updated-tool invocation shows both gains and losses across benchmarks. This may reflect an indirect influence of updated tools. Even when a tool is not invoked, its description remains in the solver's context and can affect the solver's planning and tool selection.

% \paragraph{Scope and Limitations.}
% Repeated selection on a small development set optimizes performance on the questions available during evolution. The evaluations show that the resulting operations can be reused on other tasks, but their heterogeneous outcomes indicate that development gains do not guarantee uniform improvement after transfer. Execution reliability is also important: the failed calls on VANE show that a proposed capability must operate successfully on the available video inputs before it can support reasoning. Broader development-task coverage and reliability-aware candidate evaluation are natural directions for extending the framework.

\section{Conclusion}
\label{sec:conclusion}

We presented EvoVAU, a framework that couples a frozen video solver with an external proposer to evolve reusable tools for video anomaly understanding. Cross-Version Contrastive Feedback combines current errors with historical corrections and regressions to reveal unmet analytical demands and guide tool refinement. Adoption-Aware Selection accepts a candidate only when overall development accuracy improves and corrections outnumber regressions on questions invoking refined tools. Experiments demonstrate strong in-domain and cross-domain VAU performance. Tool-use analyses further show that the solver draws on both new and existing tools according to the demands of different anomaly domains. These results demonstrate that tool evolution is an effective adaptation strategy for VAU, supporting flexible tool use and generalization across anomaly domains.

\section*{LLM Usage Statement}

We used LLMs to assist with manuscript polishing, revision, and refinement of the human-written initial tool code. LLM-based components of EvoVAU, including the frozen solvers and external proposer, are described in Sections~\ref{sec:method} and~\ref{sec:experiments}. The authors have reviewed all the LLM-assisted material and take full responsibility for the final manuscript, figures, code, and scientific claims.

{
    \small
    \bibliographystyle{ieeenat_fullname}
    \bibliography{main}
}

\clearpage
\setcounter{page}{1}
\maketitlesupplementary
\appendix 
\section{Datasets and Evaluation}
\label{sec:supp_datasets}

\subsection{Benchmark Overview}
\paragraph{VAU-Bench.}
VAU-Bench~\cite{vau-r1-arxiv2505} draws on videos from UCF-Crime~\cite{ucf-cvpr2018}, MSAD~\cite{msad-neurips2024}, and ECVA~\cite{ecva-ijcv2026}. Its multiple-choice questions ask what happened in a clip, which behavior indicates an anomaly, and which explanation best fits the events in the video. The benchmark also provides event descriptions, temporal boundaries, and reasoning annotations. We use the multiple-choice question-answering annotations of training split to construct our development set. On the test split, we report event-specific question answering (QA), anomaly classification (CLS), and binary classification accuracy (Binary).

\paragraph{VANE-Bench.}
VANE-Bench~\cite{vane-naacl2025} contains real surveillance footage and a separate collection of clips produced by five text-to-video models. The real videos come from CUHK Avenue~\cite{cuhk-avnue-iccv2013}, UCF-Crime~\cite{ucf-cvpr2018}, and UCSD Ped1 and Ped2~\cite{ucsd-ped-tpami2014}. They cover events like theft and context-dependent anomalies such as a cyclist entering a pedestrian walkway. In the generated clips, the anomaly is a visual inconsistency, where an object may change shape, appear or disappear abruptly, or pass through another solid object. The multiple-choice questions ask the solver to identify or locate the anomaly. We report accuracy separately for real and generated clips.

\paragraph{VRU-Accident.}
VRU-Accident~\cite{vru-iccvw2025} contains dashcam footage of crashes involving vulnerable road users, including pedestrians and cyclists. We evaluate six multiple-choice questions per clip, each with four answer options. Three questions cover weather and lighting, the traffic environment, and the road layout. The other three ask about the type of crash, its cause, and a possible preventive measure. We report accuracy separately for CAP~\cite{cap-mm-au-cvpr2024}, DADA-2000~\cite{dada-2000-itsc2019}, DoTA~\cite{dota-tpami2023}, and footage collected by the benchmark authors (Manual).

\paragraph{VALU.}
VALU~\cite{valu-acl2026} draws on videos from UCF-Crime~\cite{ucf-cvpr2018}, MSAD~\cite{msad-neurips2024}, and ECVA~\cite{ecva-ijcv2026}. Its five levels of annotation describe the event, the segment when the anomaly is visible, the action itself, its consequences, and the responses of those involved. We evaluate its multiple-choice Anomaly Description Choice task. For an abnormal video, the solver must select the description that matches the footage or indicate that an anomaly occurred but none of the descriptions fits. For a normal video, it must reject the proposed anomalies. We report accuracy for each of the three video sources.

\begin{table}[t]
    \centering
    \caption{Videos and questions used for tool evolution and evaluation. Videos are counted once per benchmark total; the QA and CLS rows refer to the same VAU-Bench videos.}
    \label{tab:supp_dataset_stats}
    \small
    \setlength{\tabcolsep}{8pt}
    \begin{tabular}{l l c c}
        \toprule
        Benchmark & Subset & Videos & Questions \\
        \midrule
        \multicolumn{4}{l}{\textit{Development set}} \\
        
        \multirow{4}{*}{VAU-Bench} & All & 200 & 200 \\
                  & ECVA & 99 & 99 \\
                  & UCF-Crime & 63 & 63 \\
                  & MSAD & 38 & 38 \\
        \midrule
        \multicolumn{4}{l}{\textit{In-domain evaluation}} \\
        \multirow{3}{*}{VAU-Bench} & All & 929 & 1,858 \\
                  & QA & 929 & 929 \\
                  & CLS & 929 & 929 \\
        \midrule
        \multicolumn{4}{l}{\textit{Cross-benchmark evaluation}} \\
        \multirow{3}{*}{VANE-Bench} & All & 325 & 559 \\
                   & Real & 128 & 194 \\
                   & AI-generated & 197 & 365 \\
        \addlinespace
        \multirow{5}{*}{VRU-Accident} & All & 1,000 & 6,000 \\
                     & CAP & 287 & 1,722 \\
                     & DADA-2000 & 223 & 1,338 \\
                     & DoTA & 100 & 600 \\
                     & Manual & 390 & 2,340 \\
        \addlinespace
        \multirow{4}{*}{VALU} & All & 901 & 4,861 \\
             & UCF-Crime & 269 & 1,059 \\
             & MSAD & 358 & 1,846 \\
             & ECVA & 274 & 1,956 \\
        \bottomrule
    \end{tabular}
\end{table}

\subsection{Dataset Statistics}
As shown in Table~\ref{tab:supp_dataset_stats}, we evolve the tool library using 200 VAU-Bench training videos with one question per video. The sample includes 99 videos from ECVA, 63 from UCF-Crime, and 38 from MSAD. Of the 200 videos, 62 show normal activity and the remaining 138 cover all 19 anomaly categories.

With the evolved tool library fixed, we evaluate 13,278 questions across four benchmarks. For VAU-Bench, we use the same 929 videos for event-specific QA and 20-class anomaly classification, with one question per video for each task. We report accuracy separately for the two tasks, which together account for 1858 questions. We derive the third score, binary accuracy, from the CLS predictions by mapping all anomaly classes to a single abnormal label. On VRU-Accident, each video contributes six questions. On VALU, the 4861 questions comprise 2296 matching, 2296 exclusion, and 269 rejection questions.

\section{Tool Implementation}
\label{sec:supp_tools}

\begin{table*}[t]
    \centering
    \caption{Model, inference, tool evolution, and video indexing settings for EvoVAU.}
    \label{tab:supp_implementation_settings}
    \small
    \begin{tabular*}{\textwidth}{@{\extracolsep{\fill}}l l l@{}}
        \toprule
        Component & Setting & Value \\
        % \midrule
        % \multirow{5}{*}{Models}
        %     & Solver and tool-side visual model (main) & Qwen3.8-27B~\cite{qwen38}, frozen \\
        %     & Solver and tool-side visual model (ablations) & Qwen3.5-9B~\cite{qwen35}, frozen \\
        %     & Caption and query encoder & EmbeddingGemma~2~\cite{embeddinggemma-arxiv2610}, frozen, 768 dimensions \\
        %     & Default clip captioner & Qwen3.5-9B~\cite{qwen35} \\
        %     & Offline proposer & GPT-5.6-Sol~\cite{gpt56} via Codex~\cite{codex}, high reasoning effort \\
        \midrule
        \multirow{8}{*}{Solver}
            & Context window & 65536 tokens \\
            & Temperature &  0  \\
            & Thinking mode & disabled \\
            & Output budget & 2048 tokens per turn \\
            & Tool turns per question & Up to 8 \\
            & Seed & 42 \\
            & Multimodal embedding model & EmbeddingGemma~2 \\
            & Embedding dimension & 768 \\
        \midrule
        \multirow{2}{*}{Evolution}
            & Proposal rounds & 10 \\
            & Proposer & GPT-5.6-sol with Codex \\
        \midrule
        \multirow{4}{*}{Video cache}
            & Cached frame sampling rate & 2 fps \\
            & Cached frame resolution & Maximum height 360 pixels, aspect ratio preserved \\
            & Clip duration & 10 seconds \\
            & Frames per clip for captioning & Up to 4 \\
        \bottomrule
    \end{tabular*}
\end{table*}

\subsection{Initial Tool Library}

Our initial library $T_0$ adopts the three tools introduced by Deep Video Discovery (DVD)~\cite{dvd-neurips2025}. For each video, the tools have access to cached frames, an index of timestamped clip captions, and a registry describing subjects in the video. The frozen solver chooses which tool to invoke and supplies a query or a time range according to the evidence it needs. Tools that require model inference use the same frozen model as the solver.

\paragraph{Global Browse.}
Given a question, Global Browse retrieves up to 300 clip captions by embedding similarity and arranges them in temporal order. The model reads these captions and produces a question-specific account of the events. The tool returns this account together with the subject registry, giving the solver context for interpreting who appears in the video and how an event develops.

\paragraph{Clip Search.}
The solver describes an event it wants to locate. Clip Search embeds this description, retrieves the most similar clip captions, and returns their text and time ranges in chronological order. These intervals give the solver specific moments to revisit when the captions alone leave a question unresolved.

\paragraph{Frame Inspect.}
The solver supplies a focused question and one or more time intervals. Frame Inspect orders the intervals, samples uniformly across their combined duration, and retrieves the corresponding cached frames. In the 27B run, a call uses up to four distinct frames. The model examines them together and returns a textual observation, allowing the solver to check details such as whether two subjects came into contact during a retrieved interval.

\subsection{Evolved Tool Library}

In our main experiments, the updates accepted at rounds 2 and 8 added Transient Scan and Screen Text Read, respectively. The final library retains the implementations of all three initial tools. Each addition takes a question from the solver and selects the frames to examine.

\paragraph{Transient Scan.}
When the timing of a brief event is unknown, Transient Scan searches the cached frames for candidate moments to inspect. It resizes each frame to a $160\times90$ grayscale image. For each pair of consecutive frames, it computes a \emph{frame change score} as the mean absolute difference between corresponding pixel intensities. The tool selects the transition with the highest score first, then selects up to two more whose timestamps are at least three seconds from those already selected. For each transition, the model examines up to four consecutive original cached frames in temporal order and describes the change relevant to the question. A final text-only pass combines these observations chronologically. The solver receives the ranked transition times and scores, the timestamped observations, and their synthesis. Camera motion, cuts, or lighting changes can produce a stronger peak than the event of interest.

\paragraph{Screen Text Read.}
When a question turns on visible wording or a displayed value, Screen Text Read uses the question to retrieve up to 300 captioned intervals. It keeps the earliest retrieved interval and selects up to three others by relevance, with at least five seconds between their midpoints. From each interval it samples up to four cached frames. The model transcribes relevant words and numbers visible in those frames, preserving negation and units and translating when needed. Captions guide the search, while the transcription draws on the images themselves. A final text-only pass assesses whether the visible text directly resolves the question, provides partial evidence, supplies context only, or yields no relevant reading. The tool returns this assessment with the sampled intervals and timestamps, so the solver can judge whether it also needs visual evidence of the action.

\section{Implementation Details}
\label{sec:supp_implementation}

\subsection{Model Configuration}

Table~\ref{tab:supp_implementation_settings} summarizes the solver, evolution, and video cache settings. We use Qwen3.8-27B~\cite{qwen38} as the frozen solver in the main experiment and Qwen3.5-9B~\cite{qwen35} in the feedback and selection ablations. In each run, the same model interprets frames when a tool requests visual evidence. EmbeddingGemma~2~\cite{embeddinggemma-arxiv2610} encodes clip captions and search queries for retrieval.

\subsection{Inference and Evolution Settings}

The solver can submit an answer before reaching the turn limit. We use the same inference settings to evaluate the initial and evolved libraries on each benchmark. During offline evolution, GPT-5.6-Sol~\cite{gpt56} uses Codex~\cite{codex} to propose one candidate library per round based on development feedback. The main run lasted ten rounds. At test time, the solver used the library selected in round eight.

\subsection{Library Construction}

\paragraph{Per-Video Evidence.}
Before question answering, we cache frames at 2 fps and index each video in 10-second clips. The captioner uses up to four cached frames per clip to describe the scene and its main subjects. We store each caption with its time interval and embedding, and merge the subject descriptions into a video-level registry. The initial and evolved libraries share this index and can inspect the cached frames directly.

\paragraph{Executable Tool Library.}
Each library version contains the tool functions, their prompts, and a registry of the functions available to the solver. The registry exposes each tool's name, description, and argument schema. When the solver invokes a tool, the runtime supplies the index for the current video and executes the corresponding function. The controller compares each candidate with its parent to identify added tools and changes to existing tools' interfaces, implementations, or local dependencies. It matches these tools against the recorded calls to construct the updated-tool subset used in selection.

\subsection{Tool Evolution Process}
\label{sec:supp_evolution_algorithm}

Algorithm~\ref{alg:sup_evolution} starts with the initial tool library $T_0$ and runs for $R$ proposal rounds. At the start of round $t$, $T_t$ is the retained library and $C_t$ denotes the candidate proposed in that round. Both are evaluated with the same solver $\mathcal S_\theta$, whose parameters $\theta$ remain frozen. The development set $\mathcal D$ contains $N$ examples indexed by $i$, each consisting of a video $V_i$, a question $q_i$, and its reference answer $y_i$. For any library $T$, \textsc{Evaluate} records the predicted answer $\hat y_i(T)$ and the execution trajectory $\tau_i(T)$, including tool calls and their returned observations. The correctness indicator $c_i(T)$ is one if $\hat y_i(T)=y_i$ and zero otherwise. Its mean over all $N$ questions is the development accuracy $\operatorname{Acc}(T)$.

\paragraph{Feedback and Refinement.}
$E_t$ indexes questions answered incorrectly with $T_t$. The historical sets $\mathcal H_t^+$ and $\mathcal H_t^-$ index corrections and regressions from the preceding round. Both sets are empty in the first round. The external proposer $\mathcal P$ refines $T_t$ using feedback $F_t$, which contains sampled answers and trajectories from these three sets. Once $C_t$ has been evaluated, \textsc{Compare}$(T_t,C_t)$ pairs answers to each development question. A change from incorrect with $T_t$ to correct with $C_t$ places the question in $\mathcal H_{t+1}^+$, and the reverse change places it in $\mathcal H_{t+1}^-$. These sets provide historical feedback for the next round even if $C_t$ is rejected.

\paragraph{Adoption-Aware Selection.}
$\Delta T_t$ contains the tools created or modified in $C_t$ relative to $T_t$. For question $i$, $\operatorname{Calls}(\tau_i(C_t))$ is the set of tools the solver attempted to invoke with $C_t$. The question belongs to $U_t$ if any of these tools is in $\Delta T_t$. Within this subset, $|U_t^+|$ counts questions that change from wrong to right, and $|U_t^-|$ counts those that change from right to wrong. Each question contributes at most once, regardless of the number of calls. The candidate becomes $T_{t+1}$ only if full-set accuracy increases and $|U_t^+|>|U_t^-|$. Otherwise, $T_t$ carries over. The final retained library $T_R$ is used for test-time inference.

\newcounter{suppalgorithm}
\begin{figure}[tbp]
    \small
    \hrule
    \vspace{3pt}
    \refstepcounter{suppalgorithm}\label{alg:sup_evolution}
    \noindent\textbf{Algorithm~\thesuppalgorithm}\quad Tool evolution in EvoVAU\par
    \vspace{3pt}
    \hrule
    \vspace{3pt}
    \noindent\textbf{Input} Development set $\mathcal D$, initial library $T_0$, frozen solver $\mathcal S_\theta$, proposer $\mathcal P$, and round budget $R$.\par
    \begin{tabbing}
        \hspace{1em}\=\hspace{1em}\=\kill
        \textsc{Evaluate}$(T_0,\mathcal S_\theta,\mathcal D)$\\
        $\mathcal H_0^+ \gets \varnothing,\quad \mathcal H_0^- \gets \varnothing$\\
        \textbf{for} $t=0,\ldots,R-1$ \textbf{do}\\
        \> \emph{Cross-version contrastive feedback}\\
        \> $E_t \gets \{i \mid c_i(T_t)=0\}$\\
        \> Sample records from $E_t$, $\mathcal H_t^+$, and $\mathcal H_t^-$ to form $F_t$.\\
        \> $C_t \gets \mathcal P(T_t,F_t)$\\
        \> \textsc{Evaluate}$(C_t,\mathcal S_\theta,\mathcal D)$\\
        \> $(\mathcal H_{t+1}^+,\mathcal H_{t+1}^-) \gets \textsc{Compare}(T_t,C_t)$\\[2pt]
        \> \emph{Adoption-aware selection}\\
        \> $\Delta T_t \gets$ tools newly created or modified in $C_t$\\
        \> $U_t \gets \{i \mid \operatorname{Calls}(\tau_i(C_t))\cap\Delta T_t \ne \varnothing\}$\\
        \> $U_t^+ \gets U_t \cap \mathcal H_{t+1}^+,\quad U_t^- \gets U_t \cap \mathcal H_{t+1}^-$\\
        \> \textbf{if} $\operatorname{Acc}(C_t)>\operatorname{Acc}(T_t) \;\land\; |U_t^+|>|U_t^-|$ \textbf{then}\\
        \> \> $T_{t+1} \gets C_t$\\
        \> \textbf{else}\quad $T_{t+1} \gets T_t$\\
        \textbf{return} $T_R$ for test-time inference
    \end{tabbing}
    \hrule
\end{figure}

\subsection{Solver and Proposer Prompts}
\label{sec:supp_prompts}

\paragraph{Solver Prompt.}
For each task, the solver receives the same system instructions with the initial and evolved libraries. One instruction reads, \emph{``Use each tool's description and parameter schema to decide when and how to call it.''} The solver uses the returned evidence to decide whether to inspect the video further or submit its answer through \texttt{finish}. Tool refinement can therefore change the descriptions and interfaces available to the solver while keeping its decision instructions fixed.

\paragraph{Standing Proposer Instructions.}
The proposer's \texttt{SKILL.md} asks it to examine the current implementations alongside the development trajectories. One instruction reads, \emph{``Do not impose predetermined failure categories, tool types, or a fixed tool-calling workflow.''} It then asks, \emph{``Which questions, calls, outputs, or historical comparisons show that the current tools are insufficient?''} The proposer formulates one testable hypothesis per round and explains how the proposed operation could generalize beyond the question that motivated it.

The proposer may create or modify tools and edit their descriptions, argument schemas, and internal prompts. The solver weights, its system instructions, the video index, and the evaluator remain fixed. The instructions prohibit access to held-out test data, encoding sample-specific answers in tools, and forcing calls to a refined tool. The proposer may inspect reference answers in development feedback. Tools may not read those answers at inference.

\begin{table}[t]
    \caption{VRU-Accident accuracy (\%) of six general models (26--35B total parameters) under direct inference and EvoVAU.}
    \label{tab:supp_larger_models}
    \centering
    \small
    \setlength{\tabcolsep}{2pt}
    \begin{tabular*}{\columnwidth}{@{\extracolsep{\fill}}lrrrrr@{}}
        \toprule
        Model & CAP & DADA & DoTA & Manual & Avg. \\
        \midrule
        Qwen3.8-27B & \underline{72.24} & 67.04 & 65.67 & 69.87 & 69.50 \\
        Qwen3.6-35B-A3B & 71.31 & 63.90 & 65.83 & 68.59 & 68.05 \\
        Gemma-4-26B-A4B-it & 71.78 & 65.84 & \textbf{71.00} & \textbf{70.94} & \underline{70.05} \\
        InternVL3.5-30B-A3B & 63.53 & 56.43 & 52.83 & 57.39 & 58.48 \\
        Gemma-4-31B-it & 69.86 & \underline{68.16} & 66.33 & 68.89 & 68.75 \\
        ERNIE-4.5-VL-28B-A3B & 64.92 & 59.42 & 56.67 & 59.02 & 60.57 \\
        \midrule
        \textbf{EvoVAU} & \textbf{74.10} & \textbf{69.28} & \underline{69.50} & \underline{70.51} & \textbf{71.17} \\
        \bottomrule
    \end{tabular*}
\end{table}

\paragraph{Round-Specific Proposer Prompt.}
Each round's task message directs the proposer to the parent library's source code and the current feedback file. It specifies the active feedback and selection strategies and requests one candidate library together with its hypothesis and a list of changed tools. The proposer can run lightweight checks on development examples. The controller evaluates the submitted library and determines acceptance from the solver's recorded calls and answers, as described in Algorithm~\ref{alg:sup_evolution}.

\section{Additional Results}
\label{sec:supp_additional_results}

\subsection{Comparison with Larger General Models}
\label{sec:supp_larger_models}

Table~\ref{tab:supp_larger_models} compares EvoVAU with six larger general models on VRU-Accident~\cite{vru-iccvw2025}. We first compare the two inference settings for Qwen3.8-27B. With the tool library evolved on VAU-Bench and held fixed for evaluation, overall accuracy rises from 69.50\% under direct inference to 71.17\%. Accuracy improves on all four video sources. The largest gain is on DoTA, where accuracy rises from 65.67\% to 69.50\%.

Among the six direct-inference baselines, Gemma-4-26B-A4B-it achieves the highest overall accuracy at 70.05\%. The ranking varies by source. EvoVAU ranks first on CAP and DADA and second on DoTA and Manual. On the latter two sources, Gemma-4-26B-A4B-it leads EvoVAU by 1.50 and 0.43 percentage points, respectively.

\subsection{Qualitative Analysis of Tool Use}
\label{sec:supp_qualitative_analysis}

\begin{figure*}[t]
    \centering
    \includegraphics[width=\textwidth]{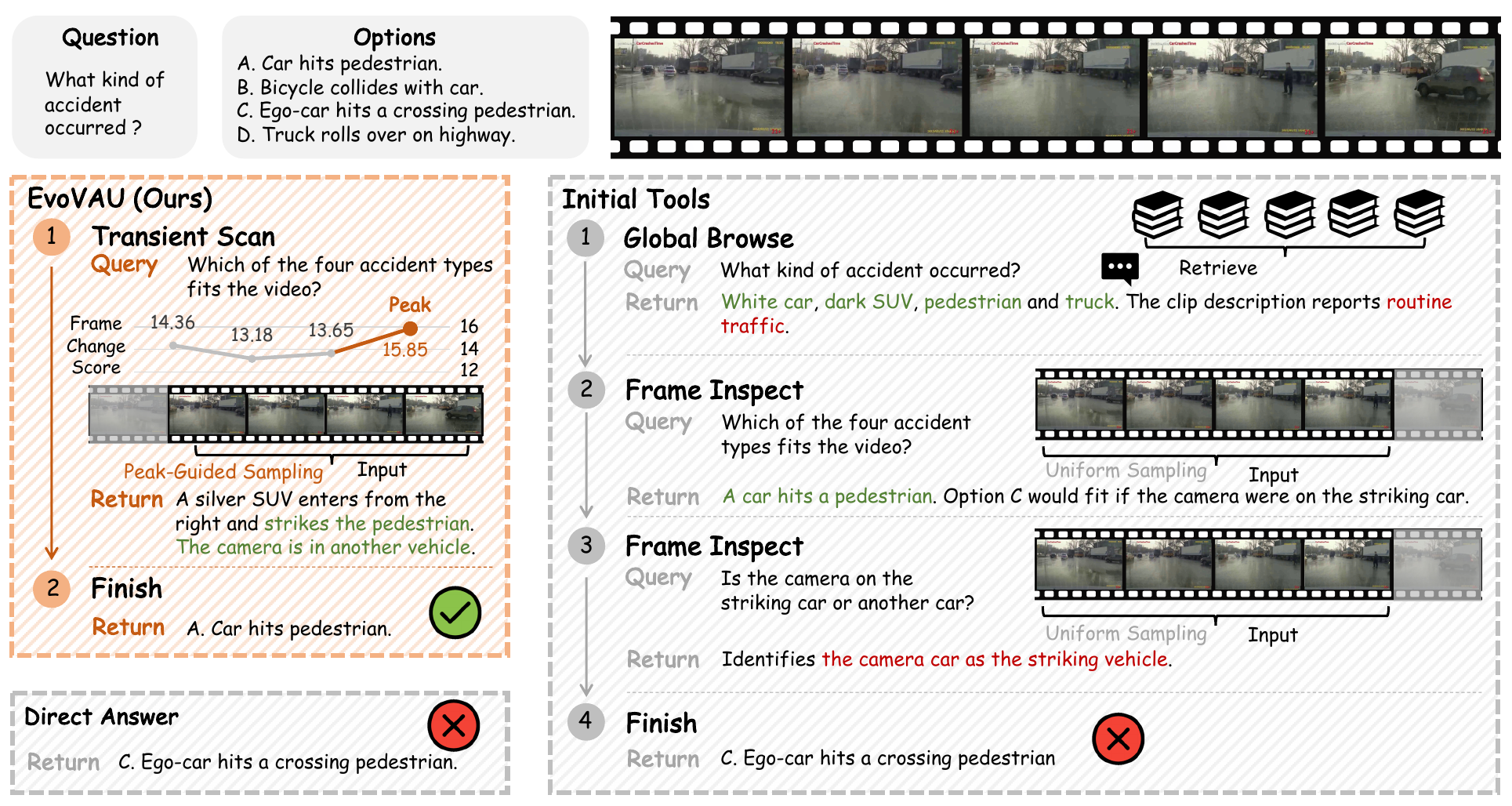}
    \caption{Accident identification on VRU-Accident with direct inference, the initial tools, and EvoVAU. In the tool responses, green highlights correct observations and red highlights errors. Faded frames were excluded from the corresponding visual inspection.}
    \label{fig:supp_vru_tool_use}
\end{figure*}

Figure~\ref{fig:supp_vru_tool_use} compares three ways of answering the same accident-type question from the CAP subset of VRU-Accident~\cite{vru-iccvw2025}. Options A and C both describe a car hitting a pedestrian. Only C specifies that the car carrying the dashcam struck the pedestrian. With direct inference, the solver selects C.

\paragraph{Initial Tools.}
Global Browse first retrieves a clip description that mentions a pedestrian crossing in front of a truck. The description omits the collision, and the tool reports routine traffic. The solver then asks Frame Inspect to compare the four answer options. The tool identifies a car approaching the pedestrian from the right and favors A, while noting that C would fit if the camera were mounted on that car.

The solver then asks whether the dashcam was mounted on the car that struck the pedestrian. Frame Inspect examines the same four uniformly sampled frames and cites the visible hood and the camera's forward motion. These cues suggest that the dashcam vehicle was moving. The tool incorrectly infers that this vehicle struck the pedestrian, and the solver submits C.

\paragraph{EvoVAU.}
The solver passes the question and its four options to Transient Scan without specifying a time range. The transition between the last two cached frames has the highest frame change score. Transient Scan therefore inspects the last four of the five cached frames, three of which were also examined by Frame Inspect. The penultimate frame shows the pedestrian in front of the truck. In the final frame, a silver SUV is visible from the side, with the pedestrian at its front. From this sequence, the tool reports that the SUV entered from the right and struck the pedestrian. The side view distinguishes the SUV from the vehicle carrying the dashcam, and the solver selects the annotated answer A after one tool call.

\end{document}